# Optimized Deployment of Unmanned Aerial Vehicles for Wildfire Detection and Monitoring

Tai Yang[1,*], Shumeng Zhang[2], Yong Wang[3], Jialei Liu[4]

## Affiliation

1. School of Mathematics and Statistics, Chongqing University, China
2. College of Computer Science, Chongqing University, China
3. College of Microelectronics and Communication Engineering, Chongqing University, China
4. School of Automation, Chongqing University, China

*corresponding: 20181973@cqu.edu.cn

## Abstract

In recent years, increased wildfires have caused irreversible damage to forest resources worldwide, threatening wildlives and human living conditions. The lack of accurate frontline information in real-time can pose great risks to firefighters. Though a plethora of machine learning algorithms have been developed to detect wildfires using aerial images and videos captured by drones, there is a lack of methods corresponding to drone deployment. We propose a wildfire rapid response system that optimizes the number and relative positions of drones to achieve full coverage of the whole wildfire area. Trained on the data from historical wildfire events, our model evaluates the possibility of wildfires at different scales and accordingly allocates the resources. It adopts plane geometry to deploy drones while balancing the capability and safety with inequality constrained nonlinear programming. The method can flexibly adapt to different terrains and the dynamic extension of the wildfire area. Lastly, the operation cost under extreme wildfire circumstances can be assessed upon the completion of the deployment. We applied our model to the wildfire data collected from eastern Victoria, Australia, and demonstrated its great potential in the real world.

## 1. Introduction

Wildfires are a big threat to ecological resources and human properties [1, 2]. Every year, a million acres of land are lost in those countries covered by large forest areas, including Australia, the United States, Canada, Brazil, and China [3-7], and the situation is increasingly severe due to the developing climate change [8, 9]. During the 2019-2020 fire season in Australia, devastating wildfires broke out in almost every state, with the most severe impact in New South Wales and eastern Victoria, resulting in at least 23 deaths, more than 1500 houses destroyed, millions of hectares of land burned, and nearly 300 million animals killed or injured [10, 11]. With the rapidly evolving global warming, it is foreseeable that the frequency and severity could increase substantially in the next century [12]. There is an urgent need to address the problem effectively.

Flame and smoke detection systems have been widely used to detect wildfire accidents and alarm the concerned authorities. The earlier sensory technologies succeeded in flame and smoke detection in relatively small spaces, such as in a building, but failed when applied in large open spaces because they require close proximity to the objects. In addition, most of them fail to accurately inform the fire's location, size, and growth rate, which are critical during wildfire fighting [13, 14]. The recent development of visual-based fire detection techniques partly alleviated the problem [15]. For example, based on random forests, Kwak et al. achieved a smoke detection rate of 93.2% and a false alarm rate of 2.2%, using features including the spatial-temporal visual patterns of intensity, wavelet energy, and motion orientation to generate feature vectors [16]. More recently, the deep learning-based wildfire detection techniques further pushed the performance. As an example, inspired by the attention mechanism [17, 18], Cao et al.'s method achieved an accuracy rate of 97.8% [19].

Conventionally, most detection systems are deployed on strategically located watchtowers equipped with cameras or sensors and satellites. However, watchtowers suffer from a limited range of view and the high cost of construction and maintenance. Though satellites have a relatively large field of view, they provide poor spatial and temporal resolution and extremely high operation costs [12, 20]. Lately, a promising surrogate, unmanned aerial vehicles (UAVs) or drones, has been proposed and quickly become popular owing to their flexibility, maneuverability, and lower operation cost. In addition, they can be easily equipped with global positioning systems (GPS), ultrahigh-resolution visual cameras, and radio communication devices. Drones are further advanced along with the improvement in portable power sources and operating software. Zhao et al. designed a specialized UAV system for wildfire monitoring and demonstrated the effectiveness of using saliency detection method and neural network in localization and recognition of wildfire based on aerial images [21]. Lee et al. developed a wildfire detection system based on deep convolution neural network (CNN) architecture utilizing UAVs, where they achieved the best accuracy of 99% by remodeling GoogLeNet [22]. These methods highlighted the promising potential of using drones in wildfire detection and monitoring.

Recently, using a group of UAVs to maximize fire coverage collaboratively has risen in popularity. A vision-based fire perimeter tracking algorithm proposed in [23] enabled the UAVs to follow the edge of a wildfire autonomously for situational awareness. The UAV team in [24] can collaboratively track a dynamic environment in the case of wildfire spreading, following the border region of the wildfire while avoiding in-flight collisions and maintaining a safe distance to fire level. Shrestha et al. used a set of potential fields to track wildfire boundaries while minimizing energy consumption, which led to 100% coverage of the boundaries by UAVs and 78.1% energy remaining on three testing scenarios [25]. Advancement in using teams of UAVs greatly expanded UAVs' utility; however, it also highlighted the need for a sophisticated method to deploy drones. Many of the existing methods in this subject only achieve full coverage of the boundaries rather than the entire wildfire area. Further, due to the dynamic nature of wildfire propagation, they may suffer from inefficient communication between the frontline and the concerned authority and incomplete fire coverage when confronting wildfires at extremely large scales. We thus were motivated to develop a model to fill these gaps.

In this work, we propose a method that comprehensively models topography, observational and communicational needs, and wildfire scale and frequency. The method maximizes the capability of drones while protecting the UAVs and firefighters to the largest extent possible. The paper is organized as follows. Section 2 specifies the preliminaries for establishing our wildfire rapid response model. Section 3 describes the architecture of the model in detail. In section 4, we show an application of our model to the target region in eastern Victoria, Australia, and lastly, section 5 concludes the paper.

## 2. Preliminary

Our overall Wildfire rapid response system includes the UAVs system, two-way radio communication, the frontline personnel, and the concerned authorities. UAVs system consists of drones carrying ultra high definition visual cameras for surveillance and situational awareness (SSA drones) and hovering drones equipped with radio repeaters located between the frontline and the relevant authorities to automatically rebroadcast signals and extend radio range (RR drones). Two-way radio communication enables the firefighters to provide status reports to the relevant authorities and allows the latter to give orders to the forward teams. The frontline personnel carries handheld two-way radios operating in the ultra-high frequency (UHF; radio frequencies ranging from 300 megahertz to 3,000 megahertz) or very high frequency (VHF; radio frequencies ranging from 30 megahertz to 300 megahertz). Their low transmitting power limits the range of handheld radios, typically a maximum of 5 watts, and is determined mainly by distance and physical topography (typically 5km in the flat region and unobstructed ground) while the weather has little effect on UHF/VHF signals. The range of the radio repeaters is also determined by distance and topography but is significantly greater than lower power handheld radios (10 watts; 20km in the flat region and unobstructed ground). Figure1 shows the overall flowchart of the whole system and some technical parameters of the UAV with types of equipment are presented in Table 1 below.

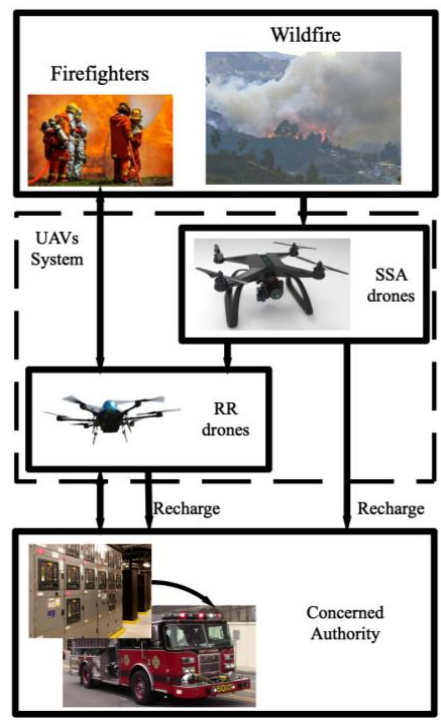

**Figure1**: overall flowchart of the wildfire rapid response system.

Table 1. Technical Parameters of the UAV.

| Technical Parameters | Value |
|---|---|
| Takeoff Weight | 17 to 24kg (full load) |
| Flight range | 30km |
| Maximum speed | 20m/s |
| Maximum flight time | 2.5hr |
| Maximum flight altitude | 4000m |
| Recharge time | 1.75hr (built-in battery) |
| cost | $10,000 |

Note: Auxiliary batteries for radios or cameras can be swapped while the built-in battery recharges.

Before applying our wildfire rapid response system to a specific area such as Victoria, Australia, it is necessary to determine the regions that deserve more attention, especially when the resources are limited. Therefore, Fire Danger Rating (FDR), Restriction in Force Time (RIFT), Fire Frequency (FF) were selected as evaluation indicators [26]. The weight of the three indicators was calculated through the method of the Analytic Hierarchy Process (AHP). After collecting the specific values of the three indicators, the method of Grey Relational Analysis (GRA) was adopted to finally determine the target regions. Let $r_i$ stand for the gray weighted correlation degree of the i-th evaluated region compared with the ideal region. It reads

$$r_i = \sum_{k=1}^{3} q_k \xi_i(k),$$

where $q_k$ is the weight of the k-th evaluation indicator, and $\xi_i(k)$ is the gray correlation degree coefficient.

Considering that one of the most crucial parts in terms of security for both firefighters and drones is the extremely high temperature, the Heat Conduction Equation was introduced to describe the evolving temperature during the process of wildfire monitoring and fighting. Let T stand for the evolving temperature, and (x, y, z) is the space coordinates in the rectangular coordinate system. It reads

$$\begin{cases} \frac{\partial T}{\partial t} - \frac{k}{C\rho}\left(\frac{\partial^2 T}{\partial x^2} + \frac{\partial^2 T}{\partial y^2} + \frac{\partial^2 T}{\partial z^2}\right) = \frac{F(x,y,z,t)}{C\rho}, \\ T|_{t=0} = \varphi(x,y,z) \end{cases}$$

where t is time, k is the thermal conductivity coefficient, C and ρ stand for the heat capacity and the density, respectively, of the circumstance, φ stands for the temperature at t=0, and F is the ignition source intensity. The analytical solution of this equation can be obtained by the Fourier transform method.

Additionally, the signal range of the handheld two-way radio communication in the UHF/VHF bands carried by the firefighters in the frontline will attenuate due to rock mass or other materials in the mountainous region. Let β stand for the attenuation coefficient. It reads

$$\beta = w\sqrt{\frac{\mu\varepsilon}{2}\left[\sqrt{1 + \frac{\sigma^2}{w^2\varepsilon^2}} - 1\right]},$$

where w represents the angular frequency of the radio, μ and σ represent the magnetic permeability and the electric conductivity of the rock, respectively, and ε denotes the dielectric coefficient of the rock.

# 3. Model and Methodology

## 3.1 Horizontal range of the radio in the mountainous region

Since our wildfire rapid response model emphasized the physical topography, which was classified into, without loss of generality, two types of terrains-flat region and mountainous region, we aimed to firstly generate the three-dimensional terrain of the target region through the method of Lagrange's interpolation based on the corresponding contour plot. For simplicity, we use cones to simulate the shape of the mountains approximately. Let H and R denote the height and the radius of the bottom of the cone, respectively (i.e., the altitude and the radius of the bottom of the mountain, correspondingly).

While operating in the flat region and unobstructed ground, the signals of both the handheld radios and the radio repeaters are stable. Let L denote the actual range of the handheld radios. Obviously, $L = 5km$ in this circumstance. However, while moving in the mountainous region, the range will drop much lower due to rock mass and other materials. Assuming that the signal spreads through homogeneous media wherever in the air or other materials, it follows from basic knowledge of analytical geometry that

$$L = \sqrt{(\sqrt{25 - z_1^2} - \frac{Rz_0}{H})^2 + (z_1 - z_0)^2 * \left[1 + (\beta - 1)\frac{2R(H - z_0)}{H\sqrt{25 - z_1^2} + R\sqrt{z_1 - 2z_0}}\right]},$$

where $(x_1, y_1, z_1)$ and $(x_0, y_0, z_0)$ denote the space coordinates of the drones and the firefighters, respectively. For RR drones, moving within such a range will suffice to ensure that the signal emitted by the handheld radios can be received.

However, the actual flight paths of the drones are strictly constrained by other additional elements such as the evolving temperature. In order to minimize the total number of RR drones as much as possible while ensuring the balance between capability and safety and other concerning factors, it is more advisable to replace the actual range of the radios mentioned above with the horizontal range. Let r denote the horizontal range, within which the radio communication between the firefighter and the drones is effective and safe enough, of the radio repeaters. This yields

$$r = \max\left(\sqrt{L^2 - z_1^2}\right);$$

$$T(x_1, y_1, z_1) \leq T_0,$$

where $T_0$ is the threshold temperature under which the drones can work efficaciously and safely. The scheme of the flyable area is shown in FigureS1.

## 3.2 Rating and Prediction of the scales of wildfires

Assuming that the wildfires start to ignite from a certain point and then extend to the surroundings with a constant speed in every direction, the wildfire area can be simplified into a circle with a constantly increasing radius. Since one of the purposes of our model is to optimize the deployment of the drones according to the scale and frequency of wildfire in the target region, it is necessary to first provide evaluation metrics to rate the wildfires at different scales, and further predict the changing likelihood of extreme wildfire accidents over the next decades. Specifically, we classified wildfires into three ratings based on the radius of the wildfire area-Rating 1 with a radius ranging from 0 to 10km, Rating 2 with a radius ranging from 10 to 40km, and Rating 3 with a radius beyond 40km. The evaluation metrics are shown in Table 2, where D represents the radius of the wildfire area. Furthermore, because the wildfire occurrence only relates to the current conditions but is not directly involved with the historical accidents in the target region, such a property prioritizes the utilization of the Markov Model to forecast the changing likelihood of extreme wildfire accidents in the future. After collecting the information of the previous wildfires with different sizes and classifying them into different ratings, we can determine the random sequence $\alpha_s$ and the corresponding state space E. After that, we can further calculate the one-step transition probability matrix P and the limit probability of every rating, where s denotes the s-th wildfire accidents that occurred in the past.

Table 2. Evaluation metrics of the scales of wildfire.

| Scale | D ≤ 10km | 10km ≤ D ≤ 40km | D ≥ 40km |
|---|---|---|---|
| Rating | 1 | 2 | 3 |

## 3.3 Optimal number and relative positions of the SSA drones

We gave priority to the number and relative positions of the SSA drones when the scale of wildfire drops within Rating 1. Let d be the horizontal range of the ultra high definition visual cameras, where valid images and videos can be captured for wildfire detection. When wildfire accidents occur, it is necessary to strategically deploy all the SSA drones to achieve complete coverage of the wildfire area for surveillance and situational awareness while minimizing the number of the SSA drones as much as possible. Therefore, the problem can be abstracted into using the fewest circles with the radius of d to completely cover the circle with the radius of D. Let n denote the minimum number of the former circles, and we have

When D ≤ d, just one SSA drone will suffice to monitor the whole wildfire area.

When D > d, we used the circular division method based on the drawer principle and the isosceles triangle method to achieve our purpose. This yields

$$n = \begin{cases} 1 & 0 < D < d \\ 3 & d < D \leq \frac{2\sqrt{3}d}{3} \\ 4 & \frac{2\sqrt{3}d}{3} < D \leq \sqrt{2}d \\ 5 & \sqrt{2}d < D \leq 2\cos\frac{\pi d}{5} \end{cases}.$$

The relative positions of the SSA drones are shown in Figure2.

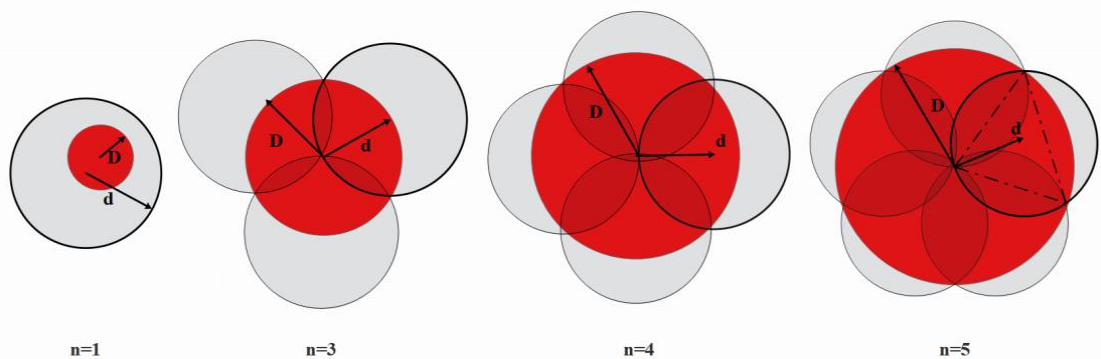

Figure2. Relative positions of the SSA drones when D ≤ 10km.

where the red and grey parts represent the wildfire area and the overall surveillance range of the

SSA drones.

However, when the value of D exceeds that of $2cos\frac{\pi d}{5}$, both the circular division method and the isosceles triangle method fail to achieve complete coverage of the wildfire area with the minimum number of the SSA drones. To address this problem, we improved our model by adopting the cellular network coverage method, which is commonly used in constructing mobile communication base stations [27, 28]. The principle of this method is, in the context of our model, replacing the circles, which have the radius of d, with their inscribed regular hexagon and then layer by layer, densely and completely, covering the circle with the radius of D. The scheme is presented in Figure3.

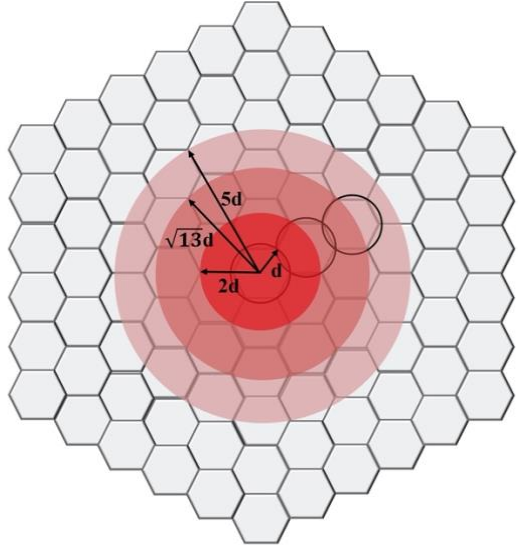

**Figure3.** Scheme of the cellular network coverage method.

where the red part still represents the wildfire area, and the grey part is a series of inscribed regular hexagons with the length of d.

Still, let n represent the minimum number of the inscribed regular hexagons (i.e., the number of the SSA drones). When $\frac{D}{d} \leq 5$, it is relatively easy to find the relationship between n and $\frac{D}{d}$ while $\frac{D}{d} > 5$, currently there is no unified formula that can precisely describe the quantitative relationship between n and $\frac{D}{d}$. Though, we can calculate the relationship approximately in an appropriate way and control the error within an acceptable range. As shown in Figure4, when the boundary of the circle with the radius of D touches a series of points such as $A_{1j}, A_{2j}$, and $B_j$, a new layer of inscribed regular hexagons must be added to ensure the dense and complete coverage of the wildfire area. If the radius D is extremely large, the distance between $A_{ij}$ and O, where O denotes the circle's center, cannot be calculated precisely by a unified formula. Instead, we replaced this distance with the distance between $M_j$ and O approximately, where $M_j$ is the midpoint of the line segment $A_{1j}A_{2j}$. Therefore, when the boundary touches those midpoints such as $M_j$, a new layer of inscribed regular hexagons is added. Let $\theta_j$ denote the angle between $OA_{1j}$ and $OM_j$. Note that:

$$\lim_{j\to\infty}\frac{|\overrightarrow{OM_j}|}{|\overrightarrow{OA_{1j}}|}=\lim_{j\to\infty}cos\theta_j=1,$$

which means that when radius D is extremely large, $OA_{1j}$ and $OM_j$ are approximately equal, and thus it is reasonable to replace $OA_{1j}$ with $OM_j$ so that we can finally conclude the formula that can determine the minimum number of SSA drones for any scale of wildfires:

$$n=\begin{cases}1 & 0<\frac{D}{d}<1\\ 3 & 1<\frac{D}{d}\leq\frac{2\sqrt{3}}{3}\\ 4 & \frac{2\sqrt{3}}{3}<\frac{D}{d}\leq\sqrt{2}\\ 5 & \sqrt{2}<\frac{D}{d}\leq 2cos\frac{\pi}{5}\\ 7 & 2cos\frac{\pi}{5}<\frac{D}{d}\leq 2\\ 19 & 2<\frac{D}{d}\leq\sqrt{13}\\ 37 & \sqrt{13}<\frac{D}{d}\leq 5\\ 1+3(a+5)(a+4) & \frac{3a+10}{2}<\frac{D}{d}\leq\frac{3a+13}{2}\end{cases},$$

where a denote non-negative integers.

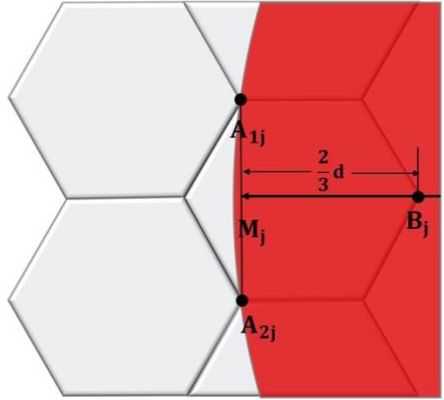

**Figure4.** Scheme of approximate calculation.

## 3.4 Optimal number and relative positions of the RR drones and the shortest deployment time

Considering that the range of the handheld two-way radios is limited in the mountainous region, the radio repeaters mounted on the RR drones are warranted to ensure efficient communication between the firefighters and the concerned authorities. Since, in the context of our model, the wildfire area extends in the shape of a circle, most of the firefighters extinguish and control the fire near the perimeter of the wildfire area. At the same time, the SSA drones keep monitoring the evolving circumstance within the boundary. Therefore, for simplicity, it is sufficient to ensure that the signal emitted from the perimeter can be received by the radio repeaters so that the problem can be abstracted into using the fewest circles with the radius of r to completely cover the periphery of the circle with the radius of D. The scheme of the relative positions of the firefighters and the RR drones is presented in FigureS2.

We aimed to minimize the number of the RR drones irregard to the radius of the wildfire area is while ensuring the radio communication is stable. Letting m denote the minimum number of the RR drones, we adopted the method of equally dividing the periphery, as shown in Figure5, when $D < \frac{r}{2}$, obviously m=1.

when $D \geq \frac{r}{2}$, it follows from the Law of Sines and the Pythagorean Theorem that

$$r^2 = \left[D^2 - \left(D \cdot \sin\frac{\pi}{m}\right)^2\right] + [D - (D \cdot \sin\frac{\pi}{m})]^2.$$

Simplifying it further, we arrived at

$$r^2 = D^2 \cdot [(\cos\frac{\pi}{m})^2 + \left(1 - \sin\frac{\pi}{m}\right)^2].$$

This yields

$$m = \begin{cases} 1 & \{D|D < \frac{r}{2}\} \\ \left\lceil \frac{\pi}{2 \cdot \arcsin\left(\frac{r}{2D}\right)} \right\rceil & \{D|D \geq \frac{r}{2}\} \cap \left\{D \left| \frac{\pi}{2 \cdot \arcsin\left(\frac{r}{2D}\right)} \in N \right.\right\}, \\ \left\lceil \frac{\pi}{2 \cdot \arcsin\left(\frac{r}{2D}\right)} \right\rceil + 1 & \{D|D \geq \frac{r}{2}\} \cap \left\{D \left| \frac{\pi}{2 \cdot \arcsin\left(\frac{r}{2D}\right)} \in (R^+\backslash N) \right.\right\} \end{cases}$$

where $R^+$ represents the set of positive real numbers and N represents the set of natural numbers.

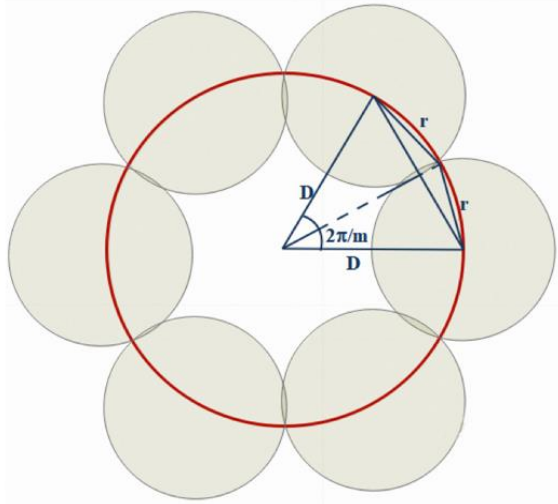

**Figure5.** Scheme of the method of periphery coverage.

In the context of our model, the RR drones always depart from the concerned authority, such as the EOC, and then fly to and hover over an appropriate position. Since we had finished the optimization of the deployment of the SSA drones, we aimed to further optimize the relative positions of the RR drones and ensure they can automatically adapt to the dynamic extension of the wildfire area, as well as minimize the time to complete the deployment of the whole UAVs system, by continuously adopting the method of equally dividing the periphery. To achieve this, firstly, the concept of feasible region was proposed to describe the horizontal flyable region of the RR drones within which they can execute the tasks efficiently without entering the wildfire area. The symbol

$C_m$ was introduced to represent the distance between one of the RR drones and the center of the wildfire area when m RR drones are deployed. Specifically, the cases $m = 1, 2, 3$ correspond to the formulas below when m=1, i.e. $0 < \frac{D}{r} \leq \frac{1}{2}$, $C_1 = r - D$, when m=2, i.e. $\frac{1}{2} < \frac{D}{r} \leq \frac{\sqrt{2}}{2}$, $C_2 = \sqrt{r^2 - D^2}$, when m=3, i.e. $\frac{\sqrt{2}}{2} < \frac{D}{r} \leq 1$, $C_3 = \frac{D}{2} + \frac{\sqrt{4r^2 - 3D^2}}{2}$, as presented in Figure6,

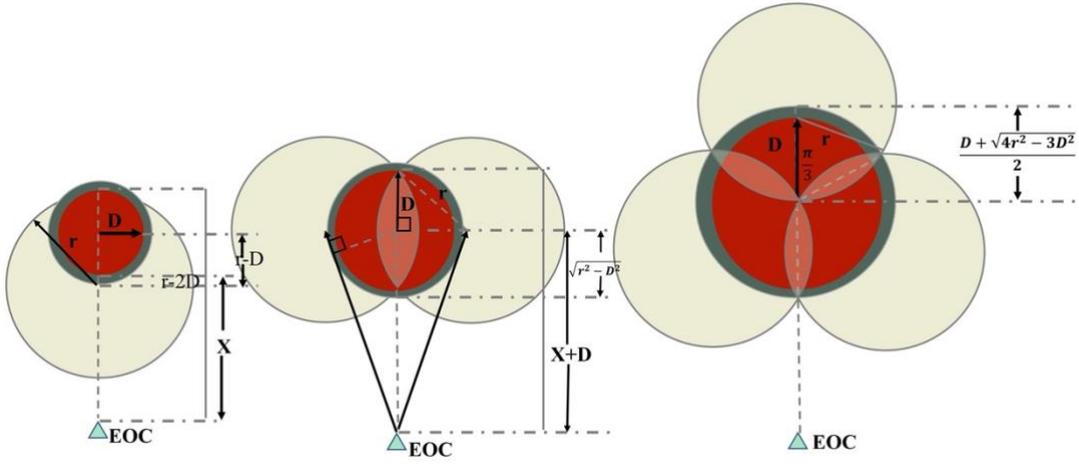

**Figure6.** scheme of the relative positions of the RR drones when $m = 1, 2, 3$.

where the blue parts stand for the feasible regions, and the light yellow parts represent the overall horizontal range of the radio repeaters, within which the radio communication is stable.

When $m \geq 4$, it follows from the Law of Cosines that for $\frac{1}{2\sin\frac{\pi}{2(m-1)}} < \frac{D}{r} \leq \frac{1}{2\sin\frac{\pi}{2m}}$,

$$\cos\frac{\pi}{m} = \frac{D^2 + C_m^2 - r^2}{2DC_m},$$

as presented in FigureS3. Simplifying the formula further, we arrived at

$$C_m^2 - \left(2D\cos\frac{\pi}{m}\right)C_m + (D^2 - r^2) = 0.$$

Note that

$$\Delta = 4D^2\cos^2\frac{\pi}{m} - 4(D^2 - r^2) = 4\left[\left(\cos^2\frac{\pi}{m} - 1\right)D^2 + r^2\right] \geq 0.$$

And thus the solution reads

$$C_m = D\cos\frac{\pi}{m} + \sqrt{\left(\cos^2\frac{\pi}{m} - 1\right)D^2 + r^2},$$

where $\frac{1}{2\sin\frac{\pi}{2(m-1)}} < \frac{D}{r} \leq \frac{1}{2\sin\frac{\pi}{2m}}$.

To ensure safety as well as flexibly adjust the relative positions during the dynamic extension of the wildfire area, the RR drones are designed to hover over the periphery of the circle with the radius of $C_m$, whose center coincides with that of the wildfire area. More specifically, each RR drone always hovers over one of the m equipartition points of the periphery when there are m RR drones deployed and their relative positions, meanwhile, remain unchanged.

Our last purpose is to minimize the time to complete the deployment of our whole UAVs system based on all our previous works, and it is sufficient to focus on the m equipartition point that is farthest from the concerned authority. We introduced the polar coordinate system in the context of our model, where the center of the wildfire area is the pole, and the symbol ρ denotes the pole axis. Let $W_m$ represent the farthest m equipartition point while $S_m$ and $Q_m$ stand for the distance between $W_m$ and the concerned authority and the minimum deployment time, respectively. Obviously, the polar coordinate of $W_m$ reads

$$\left(\frac{\pi}{m}, \quad C_m\right).$$

Assuming that the concerned authority always keeps a safety distance of b km away from the boundary of the wildfire area, we have

$$S_m = \sqrt{C_m + (D+b)^2 + 2C_m(D+b)\cos\frac{\pi}{m}},$$

with

$$S_m \leq l,$$

and

$$Q_m = \frac{S_m}{v},$$

where l and v represent the flight range and the maximum speed of the drones, respectively.

## 3.5 Additional cost caused by retirement and replacement of the drones

Extreme wildfire accidents sometimes last for even months, such as the 2019-2020 fire season in Australia [10, 11]. Long-term operation of the UAVs system under extreme wildfire circumstances will reduce the service lives of the drones because some of the components, such as the integrated circuits systems, are very sensitive to temperature. The high temperature will increase the collector current of the diode in the semiconductor devices, which will eventually lead the diode to burn out. Therefore, the cost caused by the retirement and replacement of the drones under extreme wildfire circumstances has become a factor to consider. In order to quantify this additional cost, we adopted the Poisson distribution to describe the number of the retirements per unit time approximately.

According to [29], the span of the service life of integrated circuits obeys the exponential distribution. Assume one month, without loss of generality, to be one unit time and t units time to denote the time of duration of the wildfire. Let X be a random variable that represents the service life of the drone with the probability $p_i$ describing the likelihood of retiring at the end of the i-th unit time, and $Y_i$ be another random variable that represents the number of the drones that need to be replaced at the end of the i-th unit time. It follows from the memorylessness of the exponential distribution that:

$$P\{X > t + t' | X > t\} = P\{X > t'\},$$

where $t'$ stands for the additional service life of the drone in general. This yields

$$p_i = p_{i+1}.$$

And thus

$$Y_i = Y_{i+1}.$$

Let $p = p_i = p_{i+1}$ and $Y = Y_i = Y_{i+1}$. Since the service life of each drone is independent, Y will obey the binomial distribution with parameters of $n + m$ and p

$$Y \sim B(n + m, p).$$

When the wildfire reaches an extremely large scale, the value of $n + m$ will significantly increase simultaneously. Therefore, according to Poisson's Theorem [30], Y will obey the Poisson distribution with a parameter of $(n + m)p$

$$Y \sim P((n + m)p).$$

Letting u stand for the approximate number of drones that need to be replaced at the end of every unit time, we have

$$u = [(n + m)p],$$

where $(n + m)p$ denotes the mathematical expectation of Y.

Assuming the unit price of each drone equipped with either ultra high definition visual camera or radio repeater stays constant, for simplicity, let c denote the unit price, and $G_1$ represents the additional cost caused by the retirement and replacement of the drones, and $G_0$ represents the total cost of the whole UAVs system. Obviously, we have

$$G_1 = utc.$$

And

$$G_0 = (n + m + ut)c.$$

## 4. Model Application and Results

Considering that the 2019-2020 fire season demonstrably exposed the ecosystems of Victoria, Australia to an unprecedented extent of high severity, as well as the fact that this area has a long history confronting with frequent and destructive wildfires [11], which can provide our model with a significant amount of data, we applied our wildfire rapid response system to this area following the sequence introduced in section 3.

Victoria is divided into nine fire protection zones [31]. After collecting the exact time and location of a series of historical fire events in Victoria from 1851 to 2021 [26, 31], particular regions that deserve more attention when the resources are limited were firstly filtered by our model. The values of the three indicators corresponding to each fire protection zone are presented in Table 3.

Table 3. Values of the three indicators corresponding to each fire protection zone.

| Indicator<br>Fire Division | FDR | RIFT | FF |
|---|---|---|---|
| Mallee | 11 | 198 | 3 |
| Wimmera | 10 | 170 | 7 |
| South West | 7 | 156 | 17 |
| Northern Country | 8 | 163 | 6 |
| North Central | 6 | 156 | 4 |
| North East | 6 | 157 | 15 |
| Central | 6 | 141 | 33 |

| West and South Gippsland | 5 | 141 | 45 |
| East Gippsland | 5 | 141 | 40 |

The weight of the three indicators was calculated as [0.539, 0.248, 0.213], and thus the results of the gray weighted correlation degree were reached
[$r7, r9, r8, r6, r4, r5, r3, r2, r1$] = [0.921, 0.792, 0.710, 0.600, 0.579, 0.532, 0.319, 0.142, 0.137]
Therefore, Central, East Gippsland, West and South Gippsland, and North East were selected as the target region where our UAVs system applied.

The four fire protection zones are located in eastern Victoria, with elevations ranging from sea level on the coast to 1986 meters on Mount Bogong. For simplicity, we mainly exerted our model to Mount Bogong. The simulated three-dimensional topographic map is presented in FigureS4, where H=1.986km and R=13km. Assuming that, without loss of generality, the altitude of the firefighter's position is 0.9km, and the attenuation coefficient is 0.4, we approximately calculated that r ≈ 3.3001km in the mountainous region and r ≈ 4.97km in the flat region according to the Monte Carlo Method.

Based on the Evaluation metrics of the scales of wildfire presented in Table 2, we arrived at
$$\alpha_s = \{3,1,3,2,2,3,1,3,1,3,3,1,3,2,2,2,1,2,1,3,2,3,1,1,2,2,1,1,3,3\},$$
where s = 1, 2, 3,..., 30,
and
$$E = \{1,2,3\},$$
and
$$P = \begin{pmatrix} \frac{1}{5} & \frac{1}{5} & \frac{3}{5} \\ \frac{1}{3} & \frac{4}{9} & \frac{2}{9} \\ \frac{1}{2} & \frac{3}{10} & \frac{1}{5} \end{pmatrix}.$$

Since the matrix is regular and the Markov chain is ergodic, we calculated the limit probability of every rating in the state space E as below
$$\lim_{s \to \infty} P^s = (\frac{10}{29} \quad \frac{9}{29} \quad \frac{10}{29})^T,$$

which means the probability of the occurrence of extreme wildfire accidents in the future is $\frac{10}{29}$.

Since the SSA drones and the RR drones operate during the process of wildfire detection and monitoring synchronously, we assumed that, without loss of generality, d is always equal to r. Considering that each drone has a maximum flight time of 2.5hr and a recharge time of 1.75hr, we assumed that when a drone is recharging, another drone will replace it for either surveillance and situational awareness or extending the radio range. When the scale of wildfire drops within Rating 1, the quantitative relationship between the total number of the UAVs and the value of $\frac{D}{d}$ was calculated through our model in either the flat region or the mountainous region. Specifically, when $0 < \frac{D}{d} \leq \frac{1}{2}$, n = m = 2;

when $\frac{1}{2} < \frac{D}{d} \leq \frac{\sqrt{2}}{2}$, n = 2 and m = 4;

when $\frac{\sqrt{2}}{2} < \frac{D}{d} \leq 1$, n = 2 and m = 6;

when $1 < \frac{D}{d} \leq \frac{2\sqrt{3}}{3}$, n = 6 and m = 8;

when $\frac{2\sqrt{3}}{3} < \frac{D}{d} \leq \frac{1}{2\sin\frac{\pi}{8}}$, n = m = 8;

when $\frac{1}{2\sin\frac{\pi}{8}} < \frac{D}{d} \leq \sqrt{2}$, n = 8 and m = 10;

when $\sqrt{2} < \frac{D}{d} \leq 2\cos\frac{\pi}{5}$, n = m = 10,

as shown in Figure7,

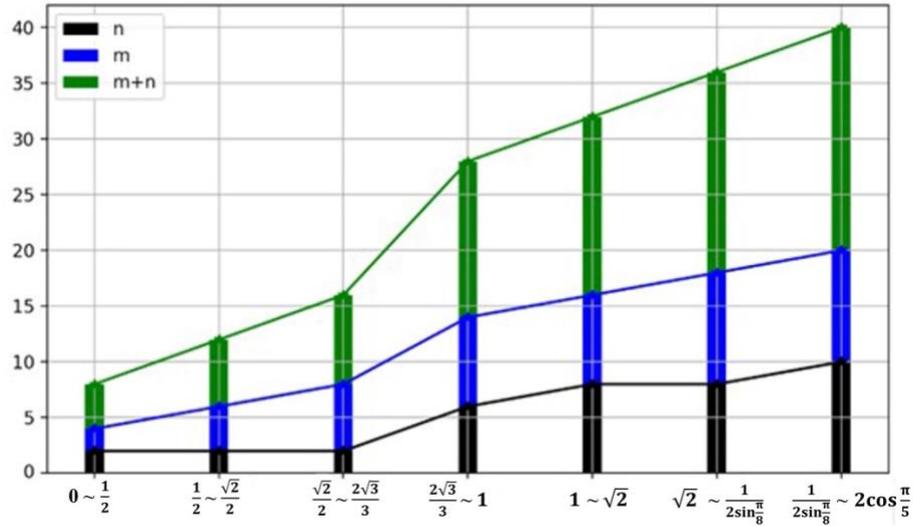

**Figure7.** Total number of the UAVs when the scale of wildfire drops within Rating 1.

where the abscissa axis represents the value of $\frac{D}{d}$, and the vertical axis represents the total number of the UAVs.

When it comes to an extreme wildfire accident, we assumed that, for simplicity, p = 0.01 and t = 12. The values of m, n, u, $G_0$, and $G_1$ corresponding to the mountainous region with the radius of the wildfire area ranging from 40km to 66km are presented in Table 4.

**Table 4:** The values of m, n, u, $G_0$, and $G_1$ under extreme wildfire circumstances in the mountainous region.

| D | m | n | u | $G_0$ | $G_1$ |
|---|---|---|---|---|---|
| 40 | 76 | 514 | 6 | 6620000 | 720000 |
| 42 | 82 | 514 | 6 | 6680000 | 720000 |
| 44 | 84 | 542 | 8 | 7220000 | 960000 |
| 46 | 88 | 572 | 8 | 7560000 | 960000 |
| 48 | 92 | 662 | 8 | 7600000 | 960000 |
| 50 | 96 | 662 | 8 | 8540000 | 960000 |
| 52 | 100 | 794 | 8 | 8580000 | 960000 |
| 54 | 104 | 794 | 10 | 10180000 | 1200000 |
| 56 | 108 | 794 | 10 | 10220000 | 1200000 |
| 58 | 112 | 794 | 10 | 10260000 | 1200000 |
| 60 | 116 | 938 | 12 | 11980000 | 1440000 |
| 62 | 120 | 938 | 12 | 12020000 | 1440000 |
| 64 | 122 | 938 | 12 | 12040000 | 1440000 |
| 66 | 126 | 1094 | 14 | 13880000 | 1680000 |

In practice, the position of the concerned authority usually keeps a safe distance away from the boundary of the wildfire area. In the context of our model, we assumed that b=5km to calculate the values of $C_m$ and $Q_m$. The quantitative relationship between $C_m$, $Q_m$, and the value of $\frac{D}{r}$ corresponding to either the flat region or the mountainous region are presented in Figure8 (a) and Figure8 (b), respectively.

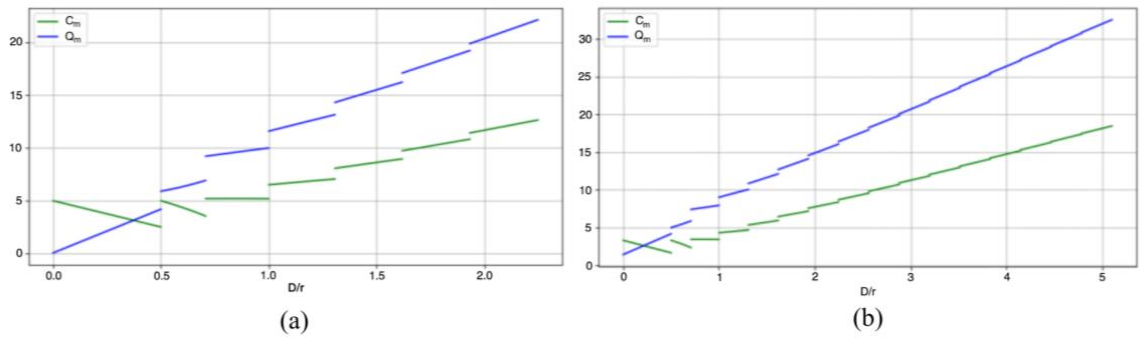

**Figure8.** Quantitative relationship between $C_m$, $Q_m$, and $\frac{D}{r}$ (a) flat region (b) mountainous region.

As can be observed from Figure8, the result of the quantitative relationship either between $C_m$ and $\frac{D}{r}$ or between $Q_m$ and $\frac{D}{r}$ is the emergence of a piecewise function, which means the optimal relative positions of the RR drones and the shortest deployment time of the UAVs system will change abruptly at some point instead of continuously during the extension of the wildfire area both in the flat region and the mountainous region. In addition, we observed a positive association between $Q_m$ and $\frac{D}{r}$, where longer deployment time is associated with a larger scale of wildfire. It is worth noting that $C_m$ piecewisely shows a negative association with $\frac{D}{r}$ from the beginning. However, this association gradually becomes positive when $\frac{D}{r} \geq 1$, which is determined by the characteristics of our method of equally dividing the periphery.

Additionally, sensitivity analysis was conducted due to several assumptions in the application of our model, such as the abstraction of the shape of mountains and the attenuation coefficient caused by the rock mass and other materials in the mountainous region.

Specifically, we exerted the sensitivity analysis to the value of $z_0$ and β by fixing the value of H to 2km. The results are presented in Figure9,

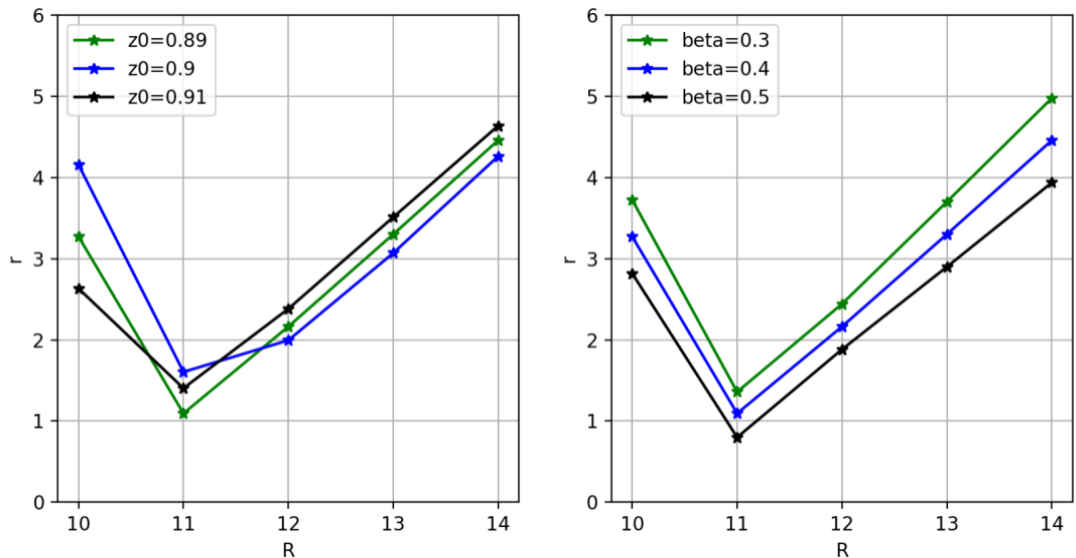

**Figure9.** Results of the sensitivity analysis.

where the abscissa axis represents the value of R, the radius of the bottom of the mountain, and the vertical axis represents the value of r, the horizontal range of the radio repeaters. It can be observed from Figure9 that our model is robust and has good stability in terms of calculating the horizontal range of the radio repeaters.

## 5. Conclusions

We proposed a novel wildfire rapid response system that achieves full coverage of the whole wildfire area for both wildfire detection and monitoring. We focused on the optimal deployment of the UAVs and how it can automatically adapt to different wildfire circumstances. The strengths of our model include its flexibility and automaticity of dealing with wildfire accidents in different physical topologies and its ability to adapt to the changing likelihood of extreme wildfire accidents in the target region. The self-adjustment of the deployment during the dynamic extension of the wildfire area ensures that the model can complete the optimal deployment of our whole UAVs system as quickly as possible. Furthermore, our model can achieve full coverage of the whole wildfire area rather than only the boundaries and the balance among capacity, safety, and economics. With the valid data of the historical wildfire events in the target region, our model can evaluate the possibility of wildfires at different scales and accordingly allocate the resource. The application of our system to the target regions in eastern Victoria, Australia showed the great potential of our model in the real world.

Though, some limitations of the current work should be recognized. First, our analysis largely relies on the historical data of the previous wildfire events in the target region, and thus it might easily be subject to data limitation, especially in those regions where the wildfire is rare. Second, our model only focuses on two types of terrains-flat region and mountainous region, but in real world, there can be a mixture of the two types. Further prospective studies with more powerful mathematical instruments dealing with the impact of complex topography on the range of radio are warranted to validate the potential adaptability of our wildfire rapid response system in the mixed type of terrains. Lastly, we assumed the wildfire area to be a circle with a constantly extending radius, which is likely to be violated when the shape of the wildfire area is almost always more irregular and even three-dimensional. These aspects shall be considered in future works.

## Conflicts of Interest

No potential conflicts of interest were disclosed.

## Authors' Contributions

**Conception and design:** Tai Yang, Shumeng Zhang, Yong Wang
**Development of methodology:** Tai Yang, Shumeng Zhang, Yong Wang
**Analysis and Interpretation of data:** Tai Yang, Shumeng Zhang, Jialei Liu
**Writing, review, and/or revision of the manuscript:** Tai Yang, Yong Wang, Jialei Liu

## Acknowledgments

We would like to thank Jianwen Xu and Chunhua Xiao for helpful discussions. We would also like to acknowledge Tao Yang for his suggestions on writing this manuscript.

## References

[1] Yuan, C., Zhang, Y., & Liu, Z. (2015). A survey on technologies for automatic forest fire monitoring, detection, and fighting using unmanned aerial vehicles and remote sensing techniques. Canadian Journal of Forest Research, 45(7), 783–792. https://doi.org/10.1139/cjfr-2014-0347

[2] Bu, F., & Gharajeh, M. S. (2019). Intelligent and vision-based fire detection systems: A survey. Image and Vision Computing, 91. https://doi.org/10.1016/j.imavis.2019.08.007

[3] Grala, K., Grala, R. K., Hussain, A., Cooke, W. H., & Varner, J. M. (2017). Impact of human factors on wildfire occurrence in Mississippi, United States. Forest Policy and Economics, 81, 38–47. https://doi.org/10.1016/j.forpol.2017.04.011

[4] Emmerton, C. A., Cooke, C. A., Hustins, S., Silins, U., Emelko, M. B., Lewis, T., Kruk, M. K., Taube, N., Zhu, D., Jackson, B., Stone, M., Kerr, J. G., & Orwin, J. F. (2020). Severe western Canadian wildfire affects water quality even at large basin scales. Water Research, 183. https://doi.org/10.1016/j.watres.2020.116071

[5] Zope, V., Dadlani, T., Matai, A., Tembhurnikar, P., & Kalani, R. (2020). IoT Sensor and Deep Neural Network based Wildfire Prediction System. 2020 4th International Conference on Intelligent Computing and Control Systems (ICICCS), 205–208. https://doi.org/10.1109/ICICCS48265.2020.9120949

[6] Wu, Z., He, H. S., Keane, R. E., Zhu, Z., Wang, Y., & Shan, Y. (2020). Current and future patterns of forest fire occurrence in China. International Journal of Wildland Fire, 29(2), 104–119. https://doi.org/10.1071/WF19039

[7] Haque, M. K., Azad, M. A. K., Hossain, M. Y., Ahmed, T., Uddin, M., & Hossain, M. M. (2021). Wildfire in Australia during 2019-2020, Its Impact on Health, Biodiversity and Environment with Some Proposals for Risk Management: A Review. Journal of Environmental Protection, 12(06), 391–414. https://doi.org/10.4236/jep.2021.126024

[8] Jolly, W. M., Cochrane, M. A., Freeborn, P. H., Holden, Z. A., Brown, T. J., Williamson, G. J., & Bowman, D. M. J. S. (2015). Climate-induced variations in global wildfire danger from 1979 to 2013. Nature Communications, 6. https://doi.org/10.1038/ncomms8537

[9] Bo, M., Mercalli, L., Pognant, F., Cat Berro, D., & Clerico, M. (2020). Urban air pollution, climate change and wildfires: The case study of an extended forest fire episode in northern Italy favoured by drought and warm weather conditions. Energy Reports, 6, 781–786. https://doi.org/10.1016/j.egyr.2019.11.002

[10] Boylan, J. L., & Lawrence, C. (2020). The development and validation of the bushfire psychological preparedness scale. International Journal of Disaster Risk Reduction, 47. https://doi.org/10.1016/j.ijdrr.2020.101530

[11] Collins, L., Bradstock, R. A., Clarke, H., Clarke, M. F., Nolan, R. H., & Penman, T. D. (2021). The 2019/2020 mega-fires exposed Australian ecosystems to an unprecedented extent of high-severity fire. In Environmental Research Letters (Vol. 16, Issue 4). IOP Publishing Ltd. https://doi.org/10.1088/1748-9326/abeb9e

[12] Hossain, F. M. A., Zhang, Y. M., & Tonima, M. A. (2020). Forest fire flame and smoke detection from UAV-captured images using fire-specific color features and multi-color space local binary pattern. Journal of Unmanned Vehicle Systems, 8(4), 285–309. https://doi.org/10.1139/juvs-2020-0009

[13] Kumar, A., Gaur, A., Singh, A., Kumar, A., Kulkarni, K. S., Lala, S., Kapoor, K., Srivastava, V., & Mukhopadhyay, S. C. (2019). Fire Sensing Technologies: A Review. In IEEE Sensors Journal (Vol. 19, Issue 9, pp. 3191–3202). Institute of Electrical and Electronics Engineers Inc. https://doi.org/10.1109/JSEN.2019.2894665

[14] Çetin, A. E., Dimitropoulos, K., Gouverneur, B., Grammalidis, N., Günay, O., Habiboğlu, Y. H., Töreyin, B. U., & Verstockt, S. (2013). Video fire detection - Review. Digital Signal Processing: A Review Journal, 23(6), 1827–1843. https://doi.org/10.1016/j.dsp.2013.07.003

[15] Gaur, A., Singh, A., Kumar, A., Kumar, A., & Kapoor, K. (2020). Video Flame and Smoke Based Fire Detection Algorithms: A Literature Review. In Fire Technology (Vol. 56, Issue 5, pp. 1943–1980). Springer. https://doi.org/10.1007/s10694-020-00986-y

[16] Kwak, J. Y., Ko, B. C., & Nam, J. Y. (2011). Forest smoke detection using CCD camera and spatial-temporal variation of smoke visual patterns. Proceedings - 2011 8th International Conference on Computer Graphics, Imaging and Visualization, CGIV 2011, 141–144. https://doi.org/10.1109/CGIV.2011.40

[17] Yang, Z., Zhang, T., & Yang, J. (2020). Research on classification algorithms for attention mechanism. 2020 19th International Symposium on Distributed Computing and Applications for Business Engineering and Science (DCABES), 194–197. https://doi.org/10.1109/DCABES50732.2020.00058

[18] Zhou, Y., & Guo, X. (2021). Small Target Segmentation Method in Complex Background Based on Attention Mechanism. 2021 IEEE International Conference on Consumer Electronics and Computer Engineering (ICCECE), 104–107. https://doi.org/10.1109/ICCECE51280.2021.9342244

[19] Cao, Y., Yang, F., Tang, Q., & Lu, X. (2019). An attention enhanced bidirectional LSTM for early forest fire smoke recognition. IEEE Access, 7, 154732–154742. https://doi.org/10.1109/ACCESS.2019.2946712

[20] Bouguettaya, A., Zarzour, H., Taberkit, A. M., & Kechida, A. (2022). A review on early wildfire detection from unmanned aerial vehicles using deep learning-based computer vision algorithms. In Signal Processing (Vol. 190). Elsevier B.V. https://doi.org/10.1016/j.sigpro.2021.108309

[21] Zhao, Y., Ma, J., Li, X., & Zhang, J. (2018). Saliency detection and deep learning-based wildfire identification in uav imagery. Sensors (Switzerland), 18(3). https://doi.org/10.3390/s18030712

[22] Lee, W., Kim, S., Lee, Y. T., Lee, H. W., & Choi, M. (2017). Deep neural networks for wild fire detection with unmanned aerial vehicle. 2017 IEEE International Conference on Consumer Electronics, ICCE 2017, 252–253. https://doi.org/10.1109/ICCE.2017.7889305

[23] Casbeer, D. W., Beard, R. W., McLain, T. W., Li, S. M., & Mehra, R. K. (2005). Forest fire monitoring with multiple small UAVs. Proceedings of the American Control Conference, 5, 3530–3535. https://doi.org/10.1109/acc.2005.1470520

[24] Pham, H. X., La, H. M., Feil-Seifer, D., & Deans, M. C. (2020). A distributed control framework of multiple unmanned aerial vehicles for dynamic wildfire tracking. IEEE Transactions on Systems, Man, and Cybernetics: Systems, 50(4), 1537–1548. https://doi.org/10.1109/TSMC.2018.2815988

[25] Shrestha, K., Dubey, R., Singandhupe, A., Louis, S., & La, H. (2021). Multi Objective UAV Network Deployment for Dynamic Fire Coverage. 2021 IEEE Congress on Evolutionary Computation (CEC), 1280–1287. https://doi.org/10.1109/cec45853.2021.9504947

[26] Tolhurst, K. G., & McCarthy, G. (2016). Effect of prescribed burning on wildfire severity: A landscape-scale case study from the 2003 fires in Victoria. Australian Forestry, 79(1), 1–14. https://doi.org/10.1080/00049158.2015.1127197

[27] Adulyasas, A., Sun, Z., & Wang, N. (2015). Connected Coverage Optimization for Sensor Scheduling in Wireless Sensor Networks. IEEE Sensors Journal, 15(7), 3877–3892. https://doi.org/10.1109/JSEN.2015.2395958

[28] Tripathi, A., Gupta, H. P., Dutta, T., Mishra, R., Shukla, K. K., & Jit, S. (2018). Coverage and Connectivity in WSNs: A Survey, Research Issues and Challenges. IEEE Access, 6, 26971–26992. https://doi.org/10.1109/ACCESS.2018.2833632

[29] Leung, Y.-W. (1993). Optimal life testing schedule for multiple types of integrated circuits. IEEE Transactions on Semiconductor Manufacturing, 6(4), 318–323. https://doi.org/10.1109/66.267641

[30] Grami, A. (2019). Basic Concepts of Probability Theory. In Probability, Random Variables, Statistics, and Random Processes: Fundamentals & Applications (pp. 1–35). Wiley. https://doi.org/10.1002/9781119300847.ch1

[31] Country Fire Authority. Fire Protection zones. https://www.cfa.vic.gov.au/warnings-restrictions/find-your-fire-district, last accessed: November 18, 2021.

# Figure Legends

**Figure1**: overall flowchart of the wildfire rapid response system.

**Figure2.** Relative positions of the SSA drones when D ≤ 10km.

**Figure3.** Scheme of the cellular network coverage method.

**Figure4.** Scheme of approximate calculation.

**Figure5.** Scheme of the method of periphery coverage.

**Figure6.** scheme of the relative positions of the RR drones when $m = 1, 2, 3$.

**Figure7.** Total number of the UAVs when the scale of wildfire drops within Rating 1.

**Figure8.** Quantitative relationship between $C_m$, $Q_m$, and $\frac{D}{r}$ (a) flat region (b) mountainous region.

**Figure9.** Results of the sensitivity analysis.

# Figures

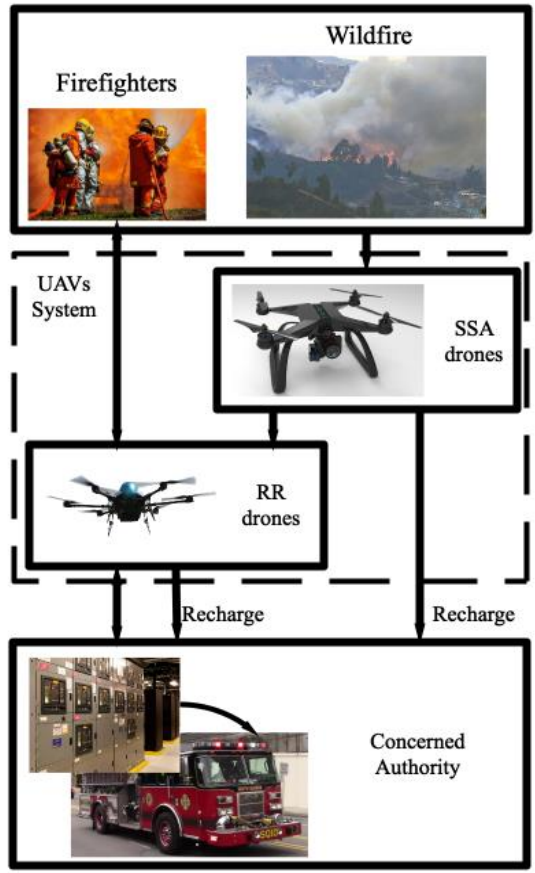

**Figure1**

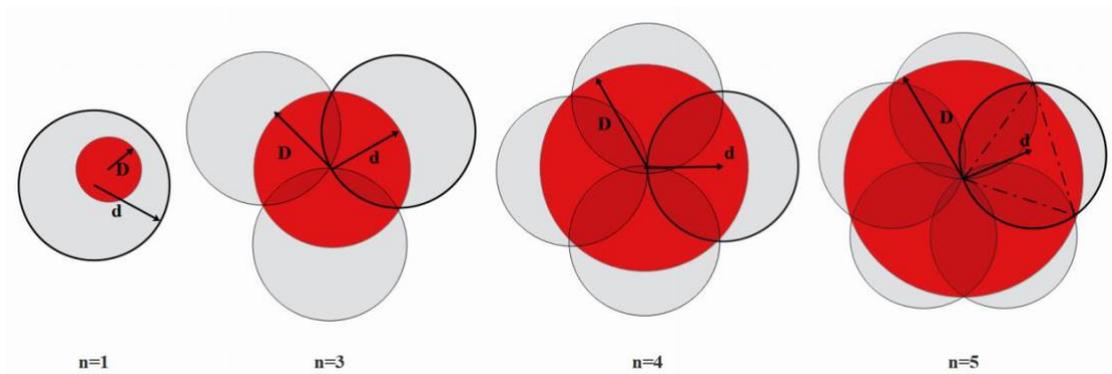

**Figure2**

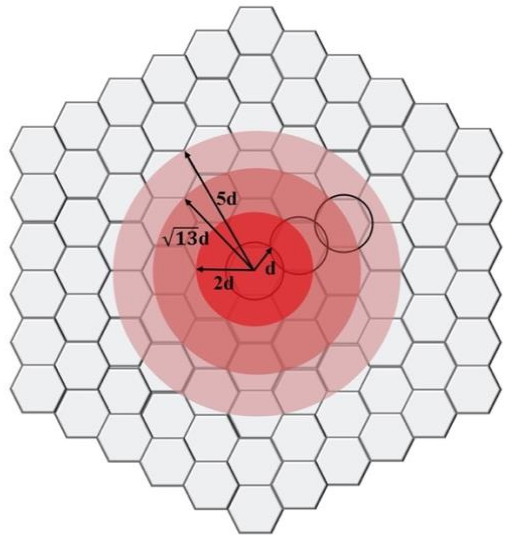

**Figure3**

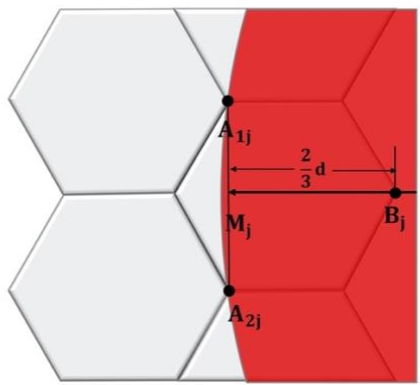

**Figure4**

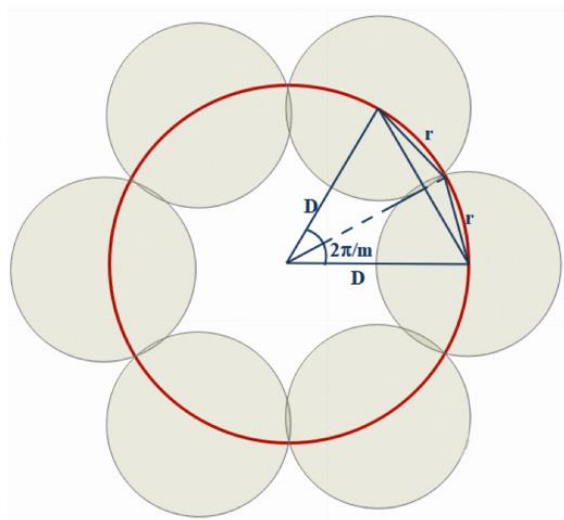

**Figure5**

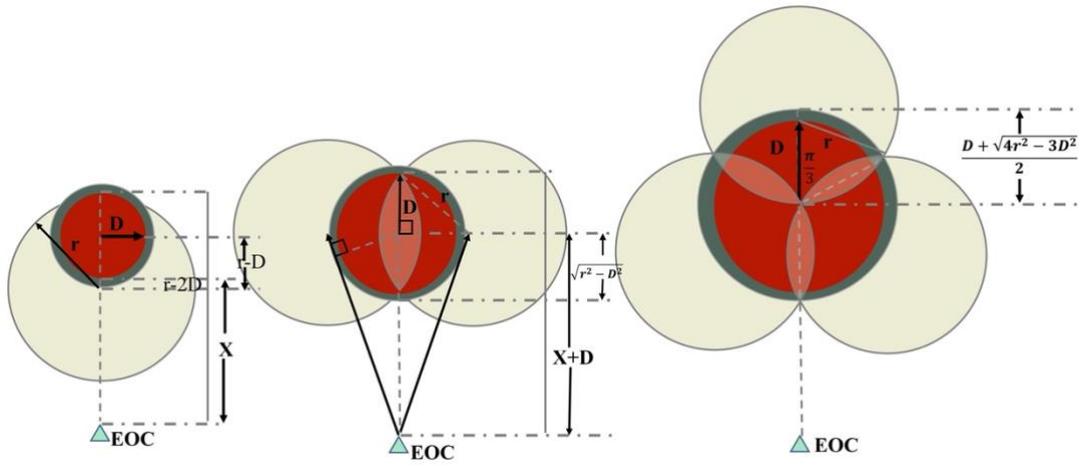

**Figure6**

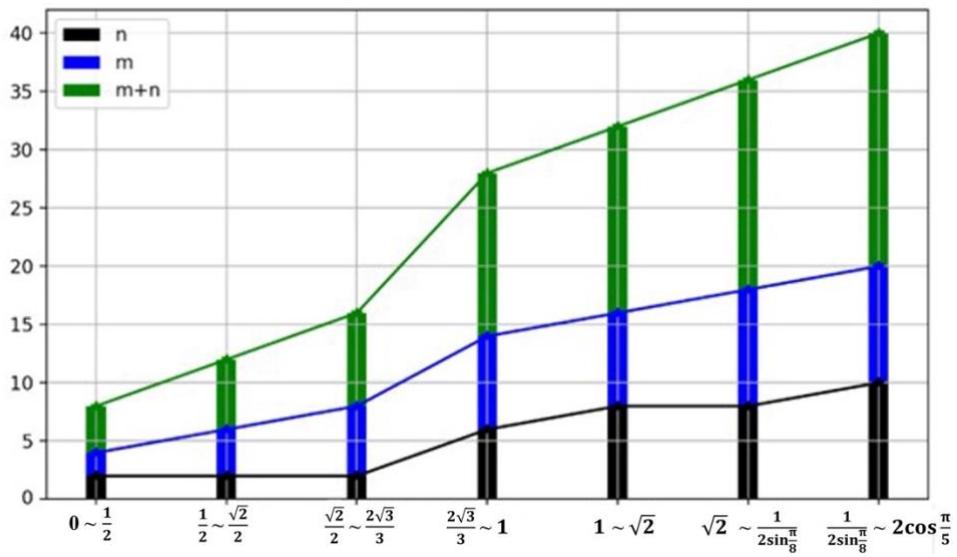

**Figure7**

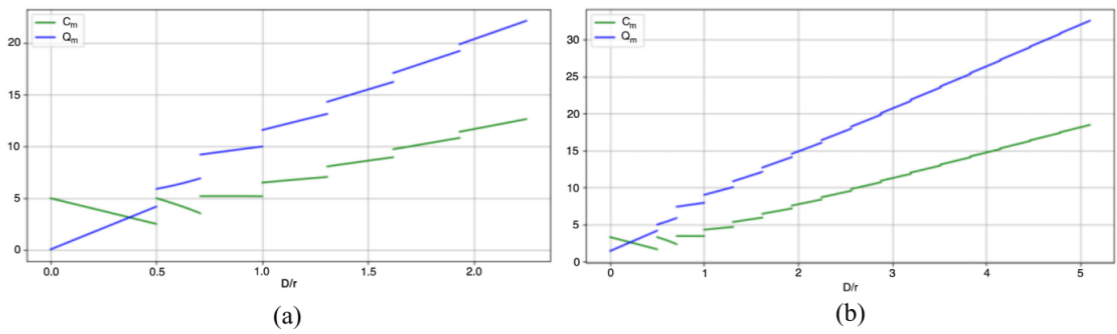

**Figure8**

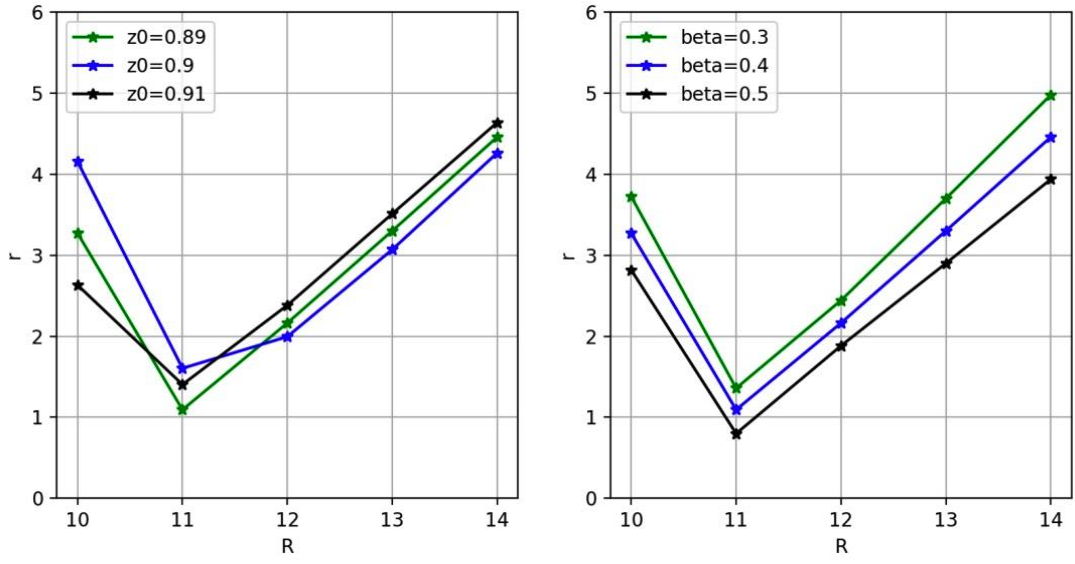

Figure9

# Supplementary Figures

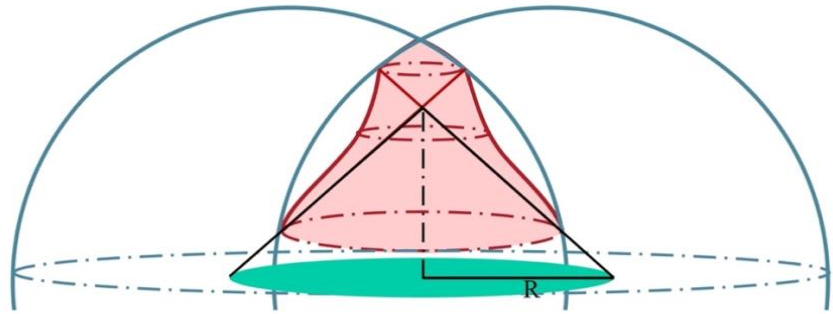

**FigureS1.** Scheme of the flyable area in the mountainous region. The mountain is simplified into a cone, and the red part represents the flyable area of the drones.

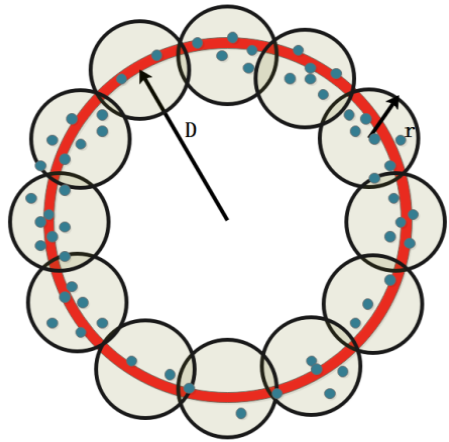

**FigureS2.** Scheme of the relative positions of the firefighters and the RR drones. The blue dots represent the positions of the firefighters, which was randomly simulated by the computer, and the light yellow parts represent the overall horizontal range of the radio repeaters, within which the radio communication is stable.

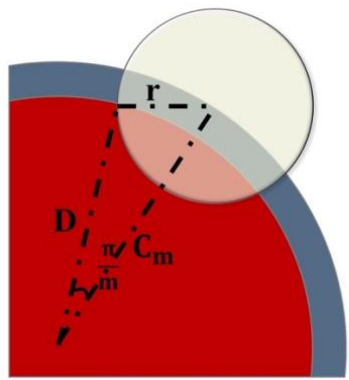

**FigureS3.** The quantitative relationship between $C_m$, $D$, and r.

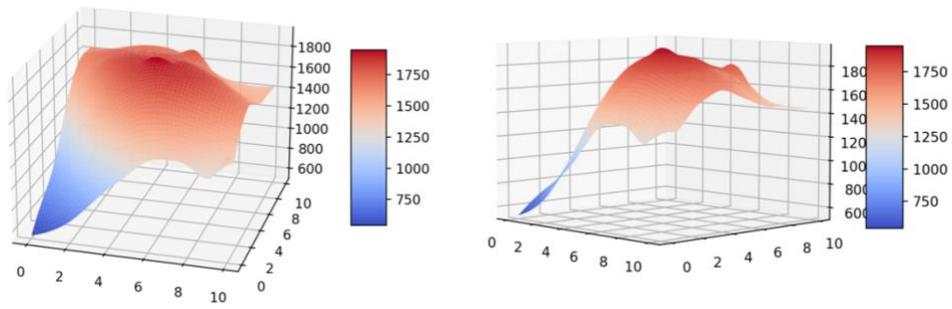

**FigureS4.** simulated three-dimensional topographic map of Mount Bogong.